\documentclass[sigconf,authorversion]{acmart}

\usepackage{graphicx}
\usepackage{amsmath}
\usepackage{amssymb}
\usepackage{booktabs}
\usepackage{array}
\usepackage{multirow}
\usepackage{color}
\usepackage{colortbl}
\usepackage{framed}
\usepackage{bm}
\usepackage{bbm}
\usepackage{xspace}
\usepackage{enumitem}
\usepackage{cite}
\usepackage{xparse}
\usepackage{xcolor}
\usepackage{algorithm}
\usepackage{lipsum}
\usepackage{listings}
\usepackage{url}

\def \ie {\emph{i.e.}}
\def \eg {\emph{e.g.}}
\def \etal {\emph{et al.}}

\newcommand{\tit}[1]{\smallbreak\noindent\textbf{#1.}}
\newcommand{\tinytit}[1]{\noindent\textbf{#1.}}

\AtBeginDocument{%
  \providecommand\BibTeX{{%
    \normalfont B\kern-0.5em{\scshape i\kern-0.25em b}\kern-0.8em\TeX}}}

\setcopyright{acmcopyright}
\copyrightyear{2022}
\acmYear{2022}
\acmDOI{XXXXXXX.XXXXXXX}

\acmConference[CBMI 2022]{International Conference on Content-based Multimedia Indexing}{Sept. 14-16, 2022}{Graz, Austria}
\acmPrice{15.00}
\acmISBN{978-1-4503-XXXX-X/18/06}

\begin{document}

\title{Retrieval-Augmented Transformer for Image Captioning}


\author{Sara Sarto} 
\affiliation{%
  \institution{University of Modena and Reggio Emilia}
  \city{Modena}
  \country{Italy}}
\email{237128@studenti.unimore.it} 

\author{Marcella Cornia}
\affiliation{%
  \institution{University of Modena and Reggio Emilia}
  \city{Modena}
  \country{Italy}}
\email{marcella.cornia@unimore.it}

\author{Lorenzo Baraldi}
\affiliation{%
  \institution{University of Modena and Reggio Emilia}
  \city{Modena}
  \country{Italy}}
\email{lorenzo.baraldi@unimore.it}

\author{Rita Cucchiara}
\affiliation{%
  \institution{University of Modena and Reggio Emilia}
  \city{Modena}
  \country{Italy}}
\email{rita.cucchiara@unimore.it}

%
%
%
%
%
%
%

\renewcommand{\shortauthors}{S. Sarto, et al.}

\begin{abstract}
Image captioning models aim at connecting Vision and Language by providing natural language descriptions of input images. In the past few years, the task has been tackled by learning parametric models and proposing visual feature extraction advancements or by modeling better multi-modal connections. In this paper, we investigate the development of an image captioning approach with a \textit{k}NN memory, with which knowledge can be retrieved from an external corpus to aid the generation process. Our architecture combines a knowledge retriever based on visual similarities, a differentiable encoder, and a \textit{k}NN-augmented attention layer to predict tokens based on the past context and on text retrieved from the external memory. Experimental results, conducted on the COCO dataset, demonstrate that employing an explicit external memory can aid the generation process and increase caption quality. Our work opens up new avenues for improving image captioning models at larger scale.
\end{abstract}

\ccsdesc[500]{Computing methodologies~Natural language generation}
\ccsdesc[500]{Computing methodologies~Visual content-based indexing and retrieval}
\ccsdesc[500]{Computing methodologies~Matching}
\ccsdesc[300]{Computing methodologies~Computer vision tasks}

\keywords{image captioning, image retrieval, vision-and-language.}


\maketitle

\section{Introduction}
\label{sec:intro}

Image captioning has recently emerged as an important task at the intersection of Computer Vision, Natural Language Processing, and Multimedia, thanks to the key role it can have to connect Vision and Language in multimedia systems~\citep{karpathy2015deep,anderson2018bottom,bigazzi2020explore}. Recent advances on image captioning, indeed, have demonstrated that fully-attentive architectures can provide high-quality image descriptions, even in few or zero-shot settings~\citep{pan2020x,li2020oscar,cornia2021universal,alayrac2022flamingo}. Regardless of this progress, accurately describing the concepts contained in input image is still a considerable challenge. While some recent works have addressed this by increasing the size of the model~\citep{alayrac2022flamingo}, thus improving its memorization capabilities, this comes at the cost of increasing the number of learnable parameters and, ultimately, the training cost.

The same issue is currently being faced in large-scale language models, where the adoption of retrieval components is being explored as a viable solution~\citep{wu2022memorizing,borgeaud2021improving}. The basic idea behind these models is that of letting the language model attend textual chunks or hidden states retrieved from an external memory, instead of relying solely on its own activations. In this manner, the memorization requirements of the model are relieved in favor of the external memory, which can handle larger-scale data and can easily be accessed via approximate nearest neighbor searches.

In this paper, we investigate the development of a retrieval component for image captioning. We propose a fully-attentive architecture with a knowledge retriever based on approximate $k$NN searches, which can provide the language model with suitable hints from an external memory. Our language model then employs a $k$NN-augmented attention layer to predict tokens based on the past context and on text retrieved from the external memory. To the best of our knowledge, our proposal is the first model to integrate a retrieval-based memory into an image captioning pipeline.

We conduct experiments on the COCO dataset for image captioning, in comparison with a fully-attentive baseline that does not employ an external memory and with an adaptation of the RETRO architecture~\citep{borgeaud2021improving}, which has been devised for large-scale language models. Our experiments show that using an external memory can significantly improve the generation quality and that adding a retrieval component to multi-modal models can be a viable solution. Further, we show that our proposed architecture overcomes the RETRO design~\citep{borgeaud2021improving} by a significant margin.
\section{Related Work}
\label{sec:related}

\tinytit{Image Captioning} Image captioning is a broad topic that has witnessed research on visual information extraction, text generation, and semantics incorporation. Over the years, different approaches have been proposed. Early works relied on CNN-based encoders and RNN-based language models~\citep{vinyals2015show,rennie2017self,donahue2015long,landi2021working}. Nowadays, attentive and Transformer-based architectures~\citep{vaswani2017attention} are often employed both in the visual encoding stage~\citep{cornia2021explaining,liu2021cptr}, either applied to image patches directly~\citep{dosovitskiy2021image,touvron2021training} or to refine features from a visual backbone, and as language models~\citep{herdade2019image,cornia2020smart,cagrandi2021learning}. Regarding language models, in the last few years, large performance improvements have come from increasing the amount of training data, model size, and performing large-scale training~\citep{Radford2019LanguageMA,brown2020language}.

The introduction of Transformer-based models in image captioning has also brought to the development of effective variants of the self-attention operator~\citep{huang2019attention,herdade2019image,pan2020x,cornia2020meshed,guo2020normalized,liu2020prophet} and to that of vision-and-language early-fusion approaches~\citep{li2020oscar,hu2021scaling,zhou2020unified} based on BERT-like architectures~\citep{devlin2018bert}. On the image encoding side, a recent paradigm is that of employing visual features extracted from large-scale multi-modal architectures~\citep{shen2021much,cornia2021universal,barraco2022unreasonable,barraco2022camel} like CLIP~\citep{radford2021learning}. 

Convolutional~\citep{aneja2018convolutional} and fully-attentive language models~\citep{luo2021dual,yang2019learning,zhang2021rstnet} based on the Transformer paradigm have been used due to the limited representation power and sequential nature of RNN-based language models and thanks to their success in NLP tasks such as machine translation and language understandings~\citep{vaswani2017attention,devlin2018bert,sukhbaatar2019augmenting}. In this paper, we follow the dominant track of employing a Transformer-based language model and propose a fully-attentive architecture augmented with retrieval abilities.

\tit{Retrieval-Augmented Approaches} Image retrieval has evolved over time from methods based on local descriptors to convolutional encoders until the use of visual Transformers~\citep{dubey2021vision,el2021training}.
Large-scale language models can perfectly memorize parts of their training data~\citep{carlini2021extracting} and increasing model size predictably improves performance on a wide range of downstream tasks~\citep{kaplan2020scaling,Radford2019LanguageMA,brown2020language}. This suggests that enhancing models with retrieval may lead to further improvements and savings in terms of model size.
In this work, we take inspiration from this line of research and investigate the incorporation of retrieval in image captioning. We conduct \textit{k}NN searches~\citep{johnson2019billion} on the extracted visual features to integrate the corresponding $k$-most similar retrieved captions with the rest of our architecture. In this way, the model can access the entire training dataset through the retrieval mechanism and is also, in principle, not limited to the data seen during training~\citep{borgeaud2021improving}. This idea allows us to go conceptually beyond the traditional Transformer language model in which the benefits of model size and data size are linked.

Retrieval-augmented language models have been recently gaining a lot of attention~\citep{guu2020realm,lewis2020retrieval,khandelwal2019generalization,izacard2020leveraging}. Most works tackling this direction have devised novel forms of attention to model the connections with the retrieved chunks of text~\citep{wu2022memorizing,yogatama2021adaptive}. 
Borgeaud~\etal~\citep{borgeaud2021improving} split input sequences into chunks, which are augmented with the $k$-nearest neighbors using a chunked cross-attention module to incorporate the retrieved text. Wu~\etal~\citep{wu2022memorizing} proposed a gated attention module to attend the internal states of a Transformer, seen during past training iterations. The usage of a learned attention gate is closely related to our formulation, even though in our case retrieval is applied towards an external memory rather than on internal activations, and we employ a single scalar gate instead of a learned per-head parameter.

\begin{figure*}[t]
    \centering
    \includegraphics[width=0.99\textwidth]{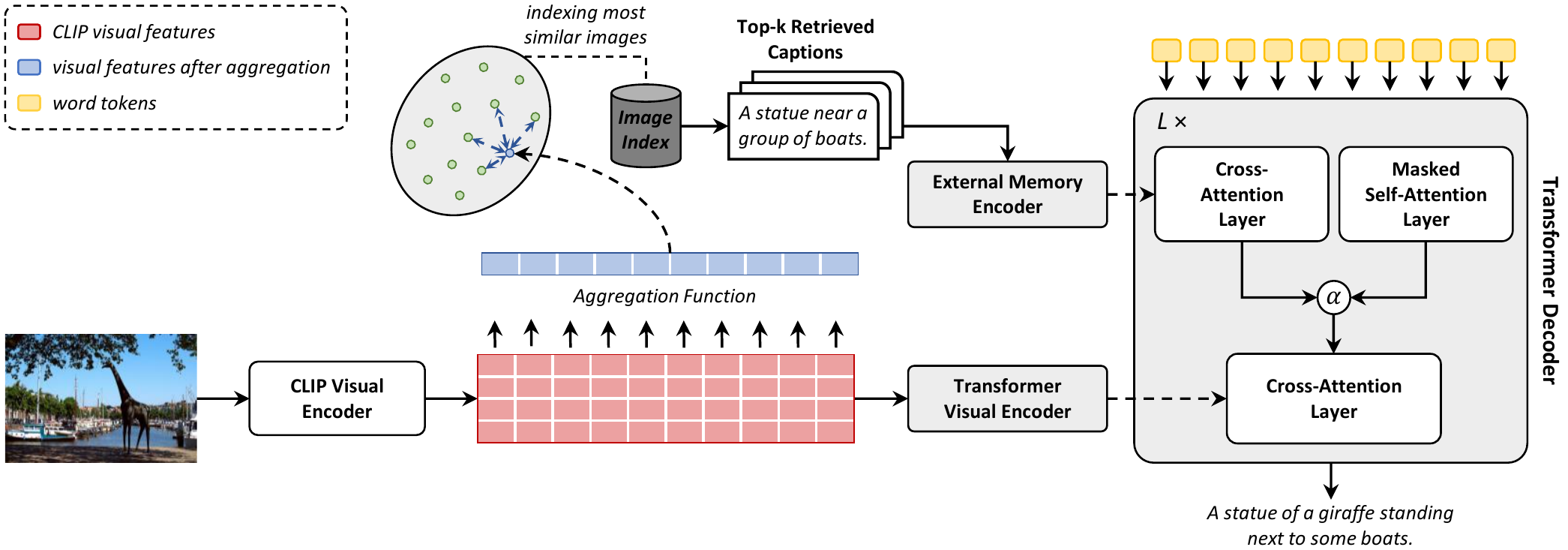}
    \caption{Illustration of our retrieval-augmented Transformer for image captioning.}
    \label{fig:architecture}
    \vspace{-0.25cm}
\end{figure*}

\section{Proposed Method}
\label{sec:method}

The goal of an image captioner is that of modeling a distribution $p(y|I)$ over possible captions $y$ given an input image $I$. During the pre-training stage, the captioner is trained with a time-wise language modeling objective: given an image $I$ and a ground-truth caption $\hat{y}$ from the training set, the objective is to predict word $\hat{y}_t$ given previous ground-truth words $\{\hat{y}_\tau\}_{\tau < t}$~\citep{cornia2020meshed,cornia2021universal}. During fine-tuning, instead, the model is usually asked to generate an entire caption $y$ without relying on previous ground-truth words. The generated caption is then usually matched with ground-truth captions to obtain a reward signal~\citep{rennie2017self}. 

Our approach decomposes the probability distribution $p(y|I)$ into two steps: \textit{retrieval} and \textit{prediction}. Firstly, given an image $I$ we retrieve possibly related descriptions $\{z_i\}_i$ from an external memory, thanks to a visual similarity space in which $k$-NN searches can be carried out. Then, we condition our language model on both the input image $I$ and the set of retrieved descriptions $\{z_i\}_i$, thus effectively modeling $p(y|\{z_i\}_i,I)$ and marginalizing over the set of retrieved captions.

\subsection{Knowledge Retriever}
\label{sec:retriever}
Given a corpus of image-text pairs and an input query image $I$, we model $p(z|I)$ by building a knowledge retrieval component. This performs an approximate $k$-nearest-neighbor search into the external memory, defined through an inner product similarity between image embeddings, \ie:
\begin{equation}
    f(I_1, I_2) = \texttt{Embed}(I_1)^\intercal \texttt{Embed}(I_2),
    \label{eq:relevance}
\end{equation}
where $\texttt{Embed}(\cdot)$ is a function that maps an image to a vector. The relevance $f(\cdot, \cdot)$ between the query image and images in the corpus is employed to sort images by decreasing similarity. Then, the knowledge retriever returns all captions associated with the selected images, as a source of conditioning for the language model.

To model the visual embedding function, we employ the visual encoder of one of the CLIP models~\citep{radford2021learning}, which have been trained contrastively to match image-text pairs. Empirically, we found this relevance function to be more robust in our scenario when compared to vision-only descriptors, as also reported in recent literature~\citep{barraco2022unreasonable,shen2021much}. While the maximum inner product search is carried out employing visual queries and values in Eq.~\ref{eq:relevance}, the search is implicitly multimodal as it happens inside a visual-semantic space. In contrast to performing a pure multi-modal search with visual queries and textual keys, however, our strategy is computationally lighter as it does not require to forward through a textual encoder.

Specifically, we select a CLIP ResNet-based~\citep{he2016deep} visual encoder. In this kind of encoder, the image is processed through a sequence of residual layers, then the grid of activations from the last convolutional layer is fed to an attention pool layer. Here, a single query is built from the global average-pooled feature vector, and all elements of the grid act as keys and values. To get a more fine-grained representation of the image and have a higher control on the pooling strategy, we directly take the grid of features from the last convolutional layer and define the $\texttt{Embed}$ function as an aggregation (\eg~average, max) of the features contained in the grid (see Fig.\ref{fig:architecture}).

\subsection{Retrieval-Augmented Language Model}
Having defined a knowledge retrieval strategy that provides captions from similar images of an external memory, we devise a retrieval-augmented language model which can predict a time-wise distribution over the vocabulary while being conditioned on both the input image and the set of retrieved captions.

The bone structure of the model is a vanilla encoder-decoder Transformer~\citep{vaswani2017attention} in which the encoder is employed to process the input image and the decoder acts as a language model. The input of the encoder consists of a flattened sequence of grid feature vectors (as described in Sec.~\ref{sec:retriever}) which are linearly projected into a vector space. The resulting vectors are then passed through a sequence of encoder layers,  each of which does dense self-attention, followed by a feed-forward network (FFN). In the encoder, attention operations are not masked, thus allowing a bidirectional encoding of input image features with complete connectivity.
The input text is instead tokenized and embedded into a vector space. The embedding vectors are then passed through a sequence of decoder layers, each of which performs a k\textit{NN-augmented attention}, a cross-attention with the last layer of the encoder, and a feed-forward network. In the decoder, we use a causal attention mask and the token embeddings of the last layer are used to predict the next token.

The \textit{k}NN-augmented attention layer combines two types of attention: like a normal self-attention layer, it attends the input subsequence encoding the past context. Plus, it performs attention over the set of retrieved captions, thus connecting to the external memory. This is modeled by first encoding all retrieved captions independently through a Transformer encoder, and then performing cross-attention over its outputs. Crucially, the same queries are employed for the self-attention over the input subsequence and for the cross-attention over the encoded retrieved captions. 

Given the input sequence of tokens $\{w_i\}_i = \{w_0, ..., w_i, ..., w_T\}$ and the set of retrieved captions $\{z_i\}_i = \{ z_0, ..., z_i, ..., z_N\}$, the \textit{k}NN-augmented attention can be written as follows:
\begin{align}
    \tilde{z}_k &= \text{MSA}(z_k, z_k) \\
    \tilde{w}^l_t &= \text{MSA}(w_t, \{w_i\}_{i=1}^t) \\
    \tilde{w}^m_t &= \text{MCA}(w_t, \{\tilde{z}_k\}_k),
\end{align}
where $k$ indicates a generic item from the set of retrieved captions, $t$ the $t$-th element of the sequence of tokens, $\{w_i\}_{i=1}^t$ is the sequence of tokens up to the $t$-th element, $\text{MSA}(x, y)$ indicates a multi-head self-attention with $x$ mapped to query and $y$ mapped to key-values, and $\text{MCA}(x, y)$ a multi-head cross-attention with $x$ as query and $y$ as key-values. The first equation refers to a self-attention between tokens of each retrieved caption, and we drop the dependency to single tokens for readability. The last equation, instead, refers to the cross-attention operation with retrieved captions. Here, given $w_t$ as query, all tokens from all retrieved captions are attended.

The outputs of the self-attention over the input subsequence and that of the cross-attention over the external memory are combined using a learned gate, which allows the model to choose between local context and retrieved captions. Formally,
\begin{equation}
    \tilde{w} = \alpha \cdot \tilde{w}^l + (1-\alpha) \tilde{w}^m,
\end{equation}
where gate $\alpha$ is learned as the sigmoid of a single scalar parameter.

As it might be noticed, gradients are not backpropagated into the external memory, which is critical to the scalability of our technique. Following a standard practice in image captioning, we first pre-train our language model with a time-wise cross-entropy loss. Then, we fine-tune using the self-critical sequence learning paradigm (SCST), which employs reinforcement learning over sampled sequences. Specifically, we employ the variant proposed in~\citep{cornia2020meshed} that employs beam search for sampling, and sets the baseline reward equal to the mean of rewards of generated captions inside a beam. We refer the reader to~\citep{rennie2017self} for a comprehensive treatment of the RL-based fine-tuning stage.

\section{Experimental Evaluation}
\label{sec:experiments}

\begin{table*}[t]
\small
\centering
\caption{Performance of the $k$ nearest-neighbor captions.}
\label{tab:simple_retrieval}
\setlength{\tabcolsep}{.3em}
\resizebox{\linewidth}{!}{
\begin{tabular}{ cc cccccc  c cccccc  c cccccc  c cccccc}
\toprule
& & \multicolumn{6}{c}{$k=5$}  & & \multicolumn{6}{c}{$k=10$} & & \multicolumn{6}{c}{$k=20$} & & \multicolumn{6}{c}{$k=40$} \\
\cmidrule{3-8} \cmidrule{10-15} \cmidrule{17-22} \cmidrule{24-29}
& & B-1 & B-4 & M & R & C & S
& & B-1 & B-4 & M & R & C & S
& & B-1 & B-4 & M & R & C & S 
& & B-1 & B-4 & M & R & C & S \\
\midrule
mean score & & 49.4 & 10.6 & 1.70 & 36.1 & 44.1 & 12.0
     & & 49.2 & 10.4 & 16.8 & 35.9 & 43.1 & 11.8 
     & & 48.9 & 10.2 & 16.7 & 35.7 & 42.1 & 11.6
     & & 48.6 & 10.0 & 16.5 & 35.4 & 41.1 & 11.4\\

max score (Oracle) & & 65.5 & 14.4 & 24.8 & 49.4 & 77.8 & 19.1
    & & 72.3 & 22.1 & 28.5 & 55.1 & 96.5 & 22.6
    & & 77.7 & 30.8 & 31.9 & 60.2 & 114.2 & 25.4
    & & 82.0 & 39.4 & 34.9 & 64.5 & 130.4 & 28.0 \\
\bottomrule
\end{tabular}
}
\vspace{-0.2cm}
\end{table*}

\subsection{Setup}
\tinytit{Dataset} Following standard image captioning approaches~\citep{anderson2018bottom,huang2019attention,cornia2020meshed}, we train and evaluate our model on the COCO dataset~\citep{lin2014microsoft}, thus not relying on large-scale image-text datasets~\citep{cornia2021universal}. COCO is composed of more than $120,000$ images, each of them associated with 5 human-collected captions. We follow the splits defined by Karpathy~\etal~\citep{karpathy2015deep}, using $5,000$ images for both validation and testing and the rest for training. 

\tit{Metrics} According to the standard evaluation protocol, we employ the complete set of captioning metrics: BLEU~\citep{papineni2002bleu}, METEOR~\citep{banerjee2005meteor}, ROUGE~\citep{lin2004rouge},  CIDEr~\citep{vedantam2015cider}, and SPICE~\citep{spice2016}.

\tit{Retrieval Index} We build our retrieval index on the COCO training set. During training, to reduce overfitting risks, we avoid retrieving captions that belong to the current training image.
We employ approximate \textit{k}NN search rather than exact \textit{k}NN search because it significantly improves
the computational speed of our model. To this aim, we employ the Faiss library~\citep{johnson2019billion} and a graph-based HNSW index with 32 links per vertex, which has a size of 6.7 GB. For simplicity, we do not employ any vector transform (\eg~PCA) or vector quantization, although they might be employed to reduce the index size and scale to larger datasets.

\tit{Implementation Details}
To represent images, we employ CLIP-RN50$\times$16~\citep{radford2021learning} intermediate features. To represent words, of both the input subsequence and retrieved sentences, we use Byte Pair Encoding (BPE)~\citep{sennrich2016neural} with a vocabulary size of $49,408$. We use standard sinusoidal positional encodings~\citep{vaswani2017attention} to represent word positions. For efficiency, the length of the output token sequence is limited to 40 tokens. Visual features and word tokens are projected into $d$-dimensional vectors with $d=384$ and fed to our Transformer-based model, which has $L=3$ layers in both encoder and decoder with six attention heads. The external memory encoder has the same number of heads and dimensionality as the rest of the model. The gate $\alpha$ is initialized to zero at the beginning of the training.

Pre-training with cross-entropy loss is performed using the LAMB optimizer~\citep{you2019large} and following the learning rate scheduling strategy of~\citep{vaswani2017attention} with a warmup equal to 6,000 iterations and a batch size of 1,080. 
For the CIDEr-based fine-tuning, we adopt the SCST strategy~\citep{rennie2017self} sampling over the $k=5$ best sequences from a beam-search scheme, using Adam~\citep{kingma2015adam} as optimizer, a batch size equal to 80, and a fixed learning rate of $5\times10^{-6}$.

All experiments are performed by paralyzing training on two Quadro RTX-5000 GPUs, using five gradient accumulation steps during both cross-entropy pre-training and CIDEr optimization. ZeRo memory offloading~\citep{rajbhandari2020zero} and mixed-precision~\citep{micikevicius2017mixed} are used to accelerate training and save memory. 

\begin{table}[t]
\small
\centering
\caption{Performance of a base Transformer captioner and our retrieval-augmented Transformer, by varying the aggregation function, the number $k$ of retrieved sentences, and the number of layers in the external memory encoder. Results are reported after cross-entropy pre-training.}
\label{tab:aggregation_results}
\setlength{\tabcolsep}{.35em}
\resizebox{\linewidth}{!}{
\begin{tabular}{cccc cccccc}
\toprule
Aggregation Function & Layers & $k$ & & B-1 & B-4 & M & R & C & S \\
\midrule
- & - & - & & 78.1 & 38.1 & 28.5 & 58.0 & 121.6 & 21.8 \\
\midrule
$\ell_2$-norm sum & 1 & 5 & & 78.3 & 38.6 & 28.9 & 58.3 & 123.1 & 21.8 \\
$\ell_2$-norm sum & 1 & 10 & & 78.5 & 38.6 & 28.8 & 58.3 & 122.7 & 22.1 \\
$\ell_2$-norm sum & 1 & 20 & & 78.3 & 38.6 & 28.9 & 58.3 & 123.8 & 21.9 \\
$\ell_2$-norm sum & 1 & 40 & & 78.2 & 39.1 & 28.7 & 57.9 & 122.8 & 22.0 \\
\midrule
max & 1 & 5 & & 78.6 & 38.6 & 28.9 & 58.3 & 123.6 & 22.0 \\
max & 1 & 10 & & 78.3 & 38.5 & 28.9 & 58.2 & 123.8 & 22.2 \\
max & 1 & 20 & & 78.3 & 38.6 & 29.0 & 58.3 & 124.0 & 22.1 \\
max & 1 & 40 & & 78.3 & 38.3 & 28.9 & 58.3 & 123.6 & 22.0 \\
\midrule
mean & 1 & 5 & & 78.6 & 38.7 & 29.1 & 58.5 & 124.0 & 22.0 \\
mean & 1 & 10 & & 78.9 & 38.9 & 28.9 & 58.5 & 124.5 & 22.1 \\
mean & 1 & 20 & & 78.5 & 38.6 & 28.9 & 58.3 & 124.2 & 22.0 \\
mean & 1 & 40 & & 78.4 & 38.4 & 28.9 & 58.3 & 123.1 & 22.0 \\
\midrule
mean & 2 & 10 & & 78.9 & 38.8 & 28.9 & 58.3 & 124.1 & 22.0 \\
mean & 3 & 10 & & 78.7 & 39.2 & 29.0 & 58.4 & 124.3 & 22.0 \\
mean & 2 & 20 & & 78.3 & 38.1 & 28.9 & 58.1 & 123.1 & 22.0 \\
mean & 3 & 20 & & 78.3 & 38.3 & 28.8 & 58.3 & 122.6 & 22.1 \\
\bottomrule
\end{tabular}
}
\vspace{-0.25cm}
\end{table}

\subsection{Quality of Nearest Neighbor Captions}
To prove the appropriateness of using nearest neighbor captions, we first investigate their quality with respect to ground-truth captions of a given test image. Specifically, given a sample image from the test set, we retrieve the $k$ nearest captions from the training set according to our relevance function. We then compare the retrieved set with ground-truth captions by computing their mean scores as well as the scores obtained by the retrieved caption with maximum similarity with the ground-truth.

Results are reported in Table~\ref{tab:simple_retrieval}. As it can be seen, retrieving a relatively limited number of captions (\eg~$k=5$) produces a set of captions that have a significant, although low, overlap with the ground-truth. Increasing the number of retrieved captions degrades mean scores. The maximum (oracle) score, instead, reaches significantly high levels, up to 130.4 CIDEr points when retrieving $k=40$ captions.
The quality reached by the oracle caption for higher $k$ outlines that the quality of the embedding space can still be significantly improved, even though ours is based on state-of-the-art descriptors. On the other side, results confirm that retrieved captions tend to become noisy when increasing $k$, and that even for small $k$ they do not provide a completely reliable signal. The language model, therefore, will need to selectively copy content from retrieved captions, paying attention to their coherency with the actual image content. 

\subsection{Model Ablation and Analysis}

\tit{Role of Different Aggregation Functions} We then move to our full model, and first analyze the results of different aggregation functions to embed visual features and retrieve the most similar images. Specifically, we consider a standard average pooling over grid features, a max pooling, and a sum of $\ell_2$-normalized features followed by an $\ell_2$-norm of the result, which has demonstrated to be effective in previous image and video retrieval works~\citep{tolias2016particular}. Results are reported in Table~\ref{tab:aggregation_results}, after cross-entropy pre-training, in comparison with a standard Transformer-based encoder-decoder model without retrieval. We can first notice that \textit{all configurations with the external memory encoder achieve better performance than the baseline} which obtains 121.6 CIDEr points, thus demonstrating the effectiveness of our retrieval-augmented architecture. When comparing the different aggregation functions, the results show that a standard mean of grid features performs generally better than the other considered aggregation functions, also according to a different number $k$ of retrieved captions.

\begin{table}[t]
\small
\centering
\caption{Ablation study results, in comparison with RETRO.}
\label{tab:ablation}
\setlength{\tabcolsep}{.35em}
\resizebox{\linewidth}{!}{
\begin{tabular}{lccccccc}
\toprule
 & & B-1 & B-4 & M & R & C & S \\
\midrule
Ours w/ RETRO block ($C = 2$)~\citep{borgeaud2021improving}  & & 75.2 & 34.5 & 26.2 & 55.7 & 108.6 & 20.2 \\
Ours w/ RETRO block ($C = 6$)~\citep{borgeaud2021improving}  & & 74.8 & 33.9 & 26.3 & 55.8 & 106.2 & 19.9 \\
\midrule
Transformer (w/o external memory) & & 78.1 & 38.1 & 28.5 & 58.0 & 121.6 & 21.8 \\
RA-Transformer (w/o gate) & & 78.3 & 38.3 & \textbf{28.9} & 58.1 & 122.5 & 21.9 \\
\textbf{RA-Transformer}  & & \textbf{78.9} & \textbf{38.9} & \textbf{28.9} & \textbf{58.5} & \textbf{124.5} & \textbf{22.1} \\
\bottomrule
\end{tabular}
}
\vspace{-0.25cm}
\end{table}

\tit{Number of Layers and Retrieved Captions} We also evaluate the effect of changing the number $k$ of retrieved sentences and the number of layers in the external memory encoder. In particular, we evaluate the captioning results employing $k=5, 10, 20, 40$ retrieved elements. From Table~\ref{tab:aggregation_results}, it can be seen that $k=10$ and $k=20$ generally lead to the best results according to almost all evaluation metrics. Additionally, we compare the results using a different number of Transformer layers (\ie~1, 2, and 3) in the external memory encoder. The architecture with a single encoding layer obtains better performance than those with an increased number of parameters. Overall, the best performance is obtained by the model with the mean as aggregation function, $k=10$ retrieved captions, and a single Transformer layer in the external memory encoder, with a CIDEr score equal to $124.5$ points. This configuration is used in all experiments reported in the rest of the paper.

\begin{figure*}[t]
    \centering
    \includegraphics[width=\textwidth]{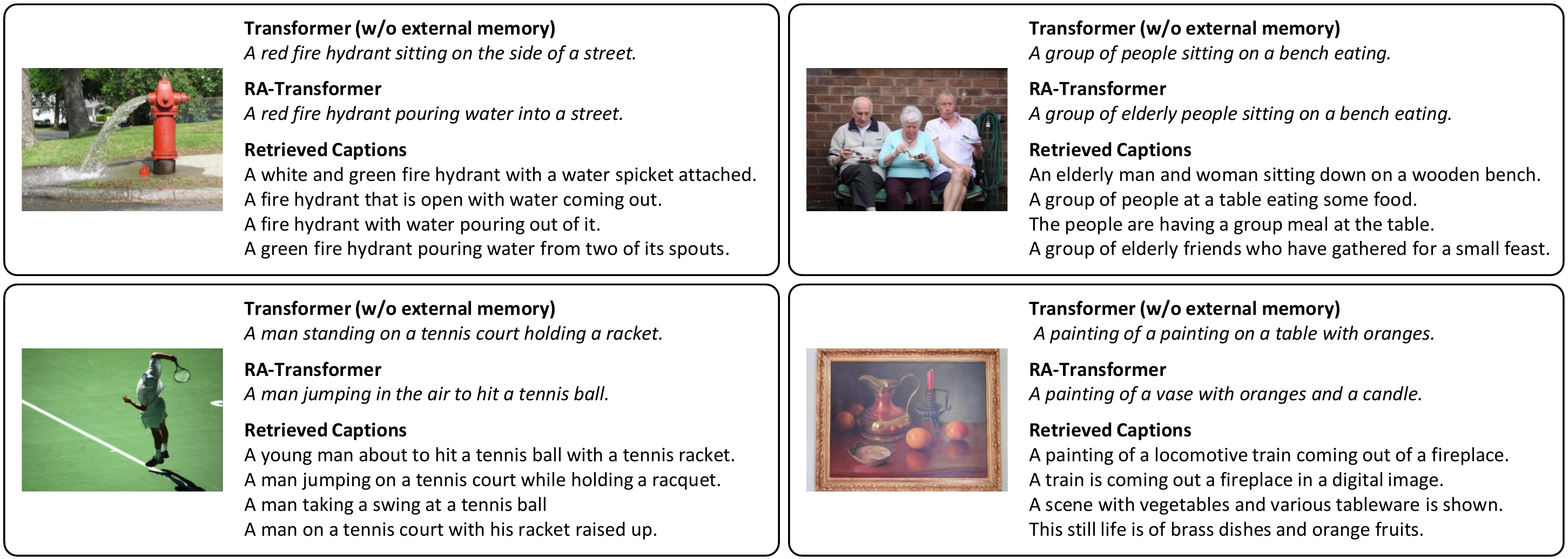}
    \caption{Qualitative results of our model, with and without the use of the external memory, with sample captions retrieved during the generation. In the bottom-right example, we show a failure case of the retrieval component.}
    \label{fig:qualitatives}
    \vspace{-0.25cm}
\end{figure*}

\tit{Retrieval-Enhanced Transformer (RETRO) Baseline} To evaluate the effectiveness of our retrieval strategy, we devise a variant of our model in which we replace the cross-attention between input tokens and retrieved captions with the chunked cross-attention mechanism proposed in~\citep{borgeaud2021improving}. We report the results in Table~\ref{tab:ablation} by varying the chunk size $C$ (\ie~$C=2$ and $C=6$). Also in this case, the results are obtained after pre-training with cross-entropy loss. When comparing the results by using different chunk sizes, it can be noticed that increasing the chunk size leads to decreasing the final results. Overall, the performance of our retrieval-augmented Transformer with multi-head cross-attention mechanism (\ie~RA-Transformer in the Table) is consistently better than that obtained by the version with chunked cross-attention.

\begin{table}[t]
\small
\centering
\caption{Comparison with state-of-the-art models on the Karpathy-test split.}
\label{tab:sota_results}
\setlength{\tabcolsep}{.35em}
\resizebox{\linewidth}{!}{
\begin{tabular}{lccccccc}
\toprule
 & & B-1 & B-4 & M & R & C & S \\
\midrule
Up-Down~\citep{anderson2018bottom} & & 79.8 & 36.3 & 27.7 & 56.9 & 120.1 & 21.4 \\
ORT~\citep{herdade2019image} & & 80.5 & 38.6 & 28.7 & 58.4 & 128.3 & 22.6 \\
GCN-LSTM~\citep{yao2018exploring} & & 80.9 & 38.3 & 28.6 & 58.5 & 128.7 & 22.1 \\
SGAE~\citep{yang2019auto} & & 81.0 & 39.0 & 28.4 & 58.9 & 129.1 & 22.2 \\ 
MT~\citep{shi2020improving} & &  80.8 & 38.9 & 28.8 & 58.7 & 129.6 & 22.3 \\
AoANet~\citep{huang2019attention} & & 80.2 & 38.9 & 29.2 & 58.8 & 129.8 & 22.4 \\
$\mathcal{M}^2$ Transformer~\citep{cornia2020meshed} & & 80.8 & 39.1 & 29.2 & 58.6 & 131.2 & 22.6 \\
X-LAN~\citep{pan2020x} & &  80.8 & 39.5 & 29.5 & 59.2 & 132.0 & 23.4 \\
X-Transformer~\citep{pan2020x} & & 80.9 & 39.7 & 29.5 & 59.1 & 132.8 & 23.4 \\
DPA~\citep{liu2020prophet} & & 80.3 & 40.5 & 29.6 & 59.2 & 133.4 & 23.3 \\
DLCT~\citep{luo2021dual} & & 81.4 & 39.8 & 29.5 & 59.1 & 133.8 & 23.0 \\
RSTNet~\citep{zhang2021rstnet} & & 81.8 & 40.1 & \textbf{29.8} & 59.5 & 135.6 & 23.3 \\
\midrule
Transformer (w/o external memory) & & 81.9 & 39.7 & 29.6 & 59.4 & 135.3 & 23.6 \\
\textbf{RA-Transformer} & & \textbf{82.4} & \textbf{40.5} & \textbf{29.8} & \textbf{59.8} & \textbf{136.5} & \textbf{23.8} \\
\bottomrule
\end{tabular}
}
\vspace{-0.25cm}
\end{table}

\tit{Role of External Memory and Learned Gate} We finally evaluate the effectiveness of both the external memory encoder and learned gate to modulate the contribution of retrieved captions. To do this, as shown in Table~\ref{tab:ablation}, we design two different baselines. In the first model, we remove the external memory encoder from our architecture thus obtaining a vanilla Transformer-based encoder-decoder model. In the second baseline, instead, we only remove the learned gating mechanism. In this case, masked self-attention and cross-attention between input tokens and retrieved captions are performed in sequence, and the result is fed to the cross-attention with visual features. 

As it can be seen, even just by introducing the external memory encoder in the architecture we can obtain enhanced results, with an improvement of 0.9 CIDEr points (\ie~121.6 vs 122.5). The final results are further improved by the introduction of the learned gate $\alpha$, with an improvement of 2 CIDEr points with respect to the architecture without gate and 2.9 CIDEr points compared to the vanilla Transformer model without external memory.

\tit{Role of Approximated $k$NN Search} We also test when employing exact $k$NN search, in place of the HNSW index. Removing the approximation in the retrieval phase brings to an increase of 0.3 CIDEr points on the COCO test set, and no significant improvement on all other metrics, thus confirming that approximate searches provide a convenient efficacy-efficiency balance when employing external memories.

\tit{Training and Inference Time Analysis} Training with cross-entropy takes around 24 hours for the model without external memory and around 30 hours for our complete model, while finetuning with reinforcement learning employs four and five days for a a standard encoder-decoder model and for our retrieval-augmented Transformer, respectively. In fact, with respect to a basic encoder-decoder captioner, our model requires running approximate $k$NN searches and introduces additional attention layers and operations. At decoding time, this entails a $26\%$ increase in execution times (from $0.170$ seconds to $0.215$ seconds to decode a single caption)\footnote{Execution times have been measured on a machine with Intel(R) Xeon(R) W-2235 CPU and Quadro RTX 5000 GPU, with a mini-batch composed of a single image and running $k$NN searches on CPU.}.  

\subsection{Comparison to the State of the Art}
We compare the results of our model with those of several recent image captioning models trained exclusively on the COCO dataset. In our analysis, we include methods with LSTM-based language models and attention over image regions such as Up-Down~\citep{anderson2018bottom}, GCN-LSTM~\citep{yao2018exploring}, SGAE~\citep{yang2019auto}, and MT~\citep{shi2020improving}, eventually enhanced with self-attention mechanisms such as AoANet~\citep{huang2019attention}, X-LAN~\citep{pan2020x}, and DPA~\citep{liu2020prophet}, and captioning architectures entirely based on fully-attentive mechanisms such as ORT~\citep{herdade2019image}, $\mathcal{M}^2$ Transformer~\citep{cornia2020meshed}, X-Transformer~\citep{pan2020x}, DLCT~\citep{luo2021dual}, and RSTNet~\citep{zhang2021rstnet}.

Table~\ref{tab:sota_results} shows the results on the Karpathy-test split after finetuning with CIDEr optimization. We report the results of our complete retrieval-augmented Transformer and of the model trained without retrieval. As it can be seen, the effectiveness of the $k$NN-augmented attention layer is confirmed also when training with reinforcement learning, with an improvement of 1.2 CIDEr points compared to the standard Transformer-based version. Additionally, we can notice that our model obtains competitive performance compared to other state-of-the-art approaches, surpassing them according to all evaluation metrics.

\subsection{Qualitative Results}
Finally, in Fig.~\ref{fig:qualitatives} we show some qualitative results generated by our model and those generated by a Transformer-based model without external memory. For each image, we also report sample captions retrieved from the external memory. As it can be seen, the majority of retrieved captions adequately match the visual content of input images and can provide a helpful external source during the generation process to improve the final results. For example, in the top-right example, we can notice that the predicted caption has very similar content to that of the retrieved sentences (\ie~``a group of elderly people''), while the model without external memory fails to generate a detailed description. Similarly, in the bottom-left example, the generated caption contains several words that also appear in the retrieved textual items (\eg~``a man jumping'' and ``a tennis ball''). This further demonstrates the effectiveness of our approach from a qualitative point of view.

In the bottom-right, we instead show an example in which the knowledge retriever partially fails to return textual sentences that effectively describe the visual content of the input image. In fact, while the image contains a painting with a vase, oranges, and other objects, some of the retrieved sentences refer to a painting with a locomotive. This confirms that additional efforts can be done to improve the quality of the retrieval embedding space and that, at the same time, the model maintains the ability to rely on input visual features when the quality of the retrieved captions is poor.

\section{Conclusion}
In this paper, we have presented a retrieval-augmented Transformer for image captioning that integrates $k$NN-augmented attention layers to generate word tokens based on textual sentences retrieved from an external memory. This enables the model to access an external corpus of textual items during the generation process thus improving the quality of generated captions. Experimental results conducted on the COCO dataset demonstrate the effectiveness of equipping a captioning architecture with retrieval abilities, opening up further research in this direction.

\section*{Acknowledgments}
\small{We thank CINECA, the Italian Supercomputing Center, for providing computational resources. This work has been supported by ``Fondazione di Modena'' and by the PRIN project ``CREATIVE: CRoss-modal understanding and gEnerATIon of Visual and tExtual content'' (CUP B87G22000460001), co-funded by the Italian Ministry of University and Research.}





\bibliographystyle{ACM-Reference-Format}
\bibliography{bibliography}


\begin{thebibliography}{65}


\ifx \showCODEN    \undefined \def \showCODEN     #1{\unskip}     \fi
\ifx \showDOI      \undefined \def \showDOI       #1{#1}\fi
\ifx \showISBNx    \undefined \def \showISBNx     #1{\unskip}     \fi
\ifx \showISBNxiii \undefined \def \showISBNxiii  #1{\unskip}     \fi
\ifx \showISSN     \undefined \def \showISSN      #1{\unskip}     \fi
\ifx \showLCCN     \undefined \def \showLCCN      #1{\unskip}     \fi
\ifx \shownote     \undefined \def \shownote      #1{#1}          \fi
\ifx \showarticletitle \undefined \def \showarticletitle #1{#1}   \fi
\ifx \showURL      \undefined \def \showURL       {\relax}        \fi
\providecommand\bibfield[2]{#2}
\providecommand\bibinfo[2]{#2}
\providecommand\natexlab[1]{#1}
\providecommand\showeprint[2][]{arXiv:#2}

\bibitem[Alayrac et~al\mbox{.}(2022)]%
        {alayrac2022flamingo}
\bibfield{author}{\bibinfo{person}{Jean-Baptiste Alayrac},
  \bibinfo{person}{Jeff Donahue}, \bibinfo{person}{Pauline Luc},
  \bibinfo{person}{Antoine Miech}, \bibinfo{person}{Iain Barr},
  \bibinfo{person}{Yana Hasson}, \bibinfo{person}{Karel Lenc},
  \bibinfo{person}{Arthur Mensch}, \bibinfo{person}{Katie Millican},
  \bibinfo{person}{Malcolm Reynolds}, {et~al\mbox{.}}}
  \bibinfo{year}{2022}\natexlab{}.
\newblock \showarticletitle{Flamingo: a Visual Language Model for Few-Shot
  Learning}.
\newblock \bibinfo{journal}{\emph{arXiv preprint arXiv:2204.14198}}
  (\bibinfo{year}{2022}).
\newblock


\bibitem[Anderson et~al\mbox{.}(2016)]%
        {spice2016}
\bibfield{author}{\bibinfo{person}{Peter Anderson}, \bibinfo{person}{Basura
  Fernando}, \bibinfo{person}{Mark Johnson}, {and} \bibinfo{person}{Stephen
  Gould}.} \bibinfo{year}{2016}\natexlab{}.
\newblock \showarticletitle{{SPICE: Semantic Propositional Image Caption
  Evaluation}}. In \bibinfo{booktitle}{\emph{ECCV}}.
\newblock


\bibitem[Anderson et~al\mbox{.}(2018)]%
        {anderson2018bottom}
\bibfield{author}{\bibinfo{person}{Peter Anderson}, \bibinfo{person}{Xiaodong
  He}, \bibinfo{person}{Chris Buehler}, \bibinfo{person}{Damien Teney},
  \bibinfo{person}{Mark Johnson}, \bibinfo{person}{Stephen Gould}, {and}
  \bibinfo{person}{Lei Zhang}.} \bibinfo{year}{2018}\natexlab{}.
\newblock \showarticletitle{{Bottom-up and top-down attention for image
  captioning and visual question answering}}. In
  \bibinfo{booktitle}{\emph{CVPR}}.
\newblock


\bibitem[Aneja et~al\mbox{.}(2018)]%
        {aneja2018convolutional}
\bibfield{author}{\bibinfo{person}{Jyoti Aneja}, \bibinfo{person}{Aditya
  Deshpande}, {and} \bibinfo{person}{Alexander~G Schwing}.}
  \bibinfo{year}{2018}\natexlab{}.
\newblock \showarticletitle{{Convolutional image captioning}}. In
  \bibinfo{booktitle}{\emph{CVPR}}.
\newblock


\bibitem[Banerjee and Lavie(2005)]%
        {banerjee2005meteor}
\bibfield{author}{\bibinfo{person}{Satanjeev Banerjee} {and}
  \bibinfo{person}{Alon Lavie}.} \bibinfo{year}{2005}\natexlab{}.
\newblock \showarticletitle{{METEOR: An automatic metric for MT evaluation with
  improved correlation with human judgments}}. In \bibinfo{booktitle}{\emph{ACL
  Workshops}}.
\newblock


\bibitem[Barraco et~al\mbox{.}(2022a)]%
        {barraco2022unreasonable}
\bibfield{author}{\bibinfo{person}{Manuele Barraco}, \bibinfo{person}{Marcella
  Cornia}, \bibinfo{person}{Silvia Cascianelli}, \bibinfo{person}{Lorenzo
  Baraldi}, {and} \bibinfo{person}{Rita Cucchiara}.}
  \bibinfo{year}{2022}\natexlab{a}.
\newblock \showarticletitle{{The Unreasonable Effectiveness of CLIP Features
  for Image Captioning: An Experimental Analysis}}. In
  \bibinfo{booktitle}{\emph{CVPR Workshops}}.
\newblock


\bibitem[Barraco et~al\mbox{.}(2022b)]%
        {barraco2022camel}
\bibfield{author}{\bibinfo{person}{Manuele Barraco}, \bibinfo{person}{Matteo
  Stefanini}, \bibinfo{person}{Marcella Cornia}, \bibinfo{person}{Silvia
  Cascianelli}, \bibinfo{person}{Lorenzo Baraldi}, {and} \bibinfo{person}{Rita
  Cucchiara}.} \bibinfo{year}{2022}\natexlab{b}.
\newblock \showarticletitle{{CaMEL: Mean Teacher Learning for Image
  Captioning}}. In \bibinfo{booktitle}{\emph{ICPR}}.
\newblock


\bibitem[Bigazzi et~al\mbox{.}(2020)]%
        {bigazzi2020explore}
\bibfield{author}{\bibinfo{person}{Roberto Bigazzi}, \bibinfo{person}{Federico
  Landi}, \bibinfo{person}{Marcella Cornia}, \bibinfo{person}{Silvia
  Cascianelli}, \bibinfo{person}{Lorenzo Baraldi}, {and} \bibinfo{person}{Rita
  Cucchiara}.} \bibinfo{year}{2020}\natexlab{}.
\newblock \showarticletitle{{Explore and Explain: Self-supervised Navigation
  and Recounting}}. In \bibinfo{booktitle}{\emph{ICPR}}.
\newblock


\bibitem[Borgeaud et~al\mbox{.}(2021)]%
        {borgeaud2021improving}
\bibfield{author}{\bibinfo{person}{Sebastian Borgeaud}, \bibinfo{person}{Arthur
  Mensch}, \bibinfo{person}{Jordan Hoffmann}, \bibinfo{person}{Trevor Cai},
  \bibinfo{person}{Eliza Rutherford}, \bibinfo{person}{Katie Millican},
  \bibinfo{person}{George van~den Driessche}, \bibinfo{person}{Jean-Baptiste
  Lespiau}, \bibinfo{person}{Bogdan Damoc}, \bibinfo{person}{Aidan Clark},
  {et~al\mbox{.}}} \bibinfo{year}{2021}\natexlab{}.
\newblock \showarticletitle{Improving language models by retrieving from
  trillions of tokens}.
\newblock \bibinfo{journal}{\emph{arXiv preprint arXiv:2112.04426}}
  (\bibinfo{year}{2021}).
\newblock


\bibitem[Brown et~al\mbox{.}(2020)]%
        {brown2020language}
\bibfield{author}{\bibinfo{person}{Tom Brown}, \bibinfo{person}{Benjamin Mann},
  \bibinfo{person}{Nick Ryder}, \bibinfo{person}{Melanie Subbiah},
  \bibinfo{person}{Jared~D Kaplan}, \bibinfo{person}{Prafulla Dhariwal},
  \bibinfo{person}{Arvind Neelakantan}, \bibinfo{person}{Pranav Shyam},
  \bibinfo{person}{Girish Sastry}, \bibinfo{person}{Amanda Askell},
  {et~al\mbox{.}}} \bibinfo{year}{2020}\natexlab{}.
\newblock \showarticletitle{Language models are few-shot learners}. In
  \bibinfo{booktitle}{\emph{NeurIPS}}.
\newblock


\bibitem[Cagrandi et~al\mbox{.}(2021)]%
        {cagrandi2021learning}
\bibfield{author}{\bibinfo{person}{Marco Cagrandi}, \bibinfo{person}{Marcella
  Cornia}, \bibinfo{person}{Matteo Stefanini}, \bibinfo{person}{Lorenzo
  Baraldi}, {and} \bibinfo{person}{Rita Cucchiara}.}
  \bibinfo{year}{2021}\natexlab{}.
\newblock \showarticletitle{{Learning to Select: A Fully Attentive Approach for
  Novel Object Captioning}}. In \bibinfo{booktitle}{\emph{ICMR}}.
\newblock


\bibitem[Carlini et~al\mbox{.}(2021)]%
        {carlini2021extracting}
\bibfield{author}{\bibinfo{person}{Nicholas Carlini}, \bibinfo{person}{Florian
  Tramer}, \bibinfo{person}{Eric Wallace}, \bibinfo{person}{Matthew Jagielski},
  \bibinfo{person}{Ariel Herbert-Voss}, \bibinfo{person}{Katherine Lee},
  \bibinfo{person}{Adam Roberts}, \bibinfo{person}{Tom Brown},
  \bibinfo{person}{Dawn Song}, \bibinfo{person}{Ulfar Erlingsson},
  {et~al\mbox{.}}} \bibinfo{year}{2021}\natexlab{}.
\newblock \showarticletitle{Extracting training data from large language
  models}. In \bibinfo{booktitle}{\emph{USENIX Security Symposium}}.
\newblock


\bibitem[Cornia et~al\mbox{.}(2020a)]%
        {cornia2020smart}
\bibfield{author}{\bibinfo{person}{Marcella Cornia}, \bibinfo{person}{Lorenzo
  Baraldi}, {and} \bibinfo{person}{Rita Cucchiara}.}
  \bibinfo{year}{2020}\natexlab{a}.
\newblock \showarticletitle{{SMArT: Training Shallow Memory-aware Transformers
  for Robotic Explainability}}. In \bibinfo{booktitle}{\emph{ICRA}}.
\newblock


\bibitem[Cornia et~al\mbox{.}(2021)]%
        {cornia2021explaining}
\bibfield{author}{\bibinfo{person}{Marcella Cornia}, \bibinfo{person}{Lorenzo
  Baraldi}, {and} \bibinfo{person}{Rita Cucchiara}.}
  \bibinfo{year}{2021}\natexlab{}.
\newblock \showarticletitle{Explaining transformer-based image captioning
  models: An empirical analysis}.
\newblock \bibinfo{journal}{\emph{AI Communications}} (\bibinfo{year}{2021}),
  \bibinfo{pages}{1--19}.
\newblock


\bibitem[Cornia et~al\mbox{.}(2022)]%
        {cornia2021universal}
\bibfield{author}{\bibinfo{person}{Marcella Cornia}, \bibinfo{person}{Lorenzo
  Baraldi}, \bibinfo{person}{Giuseppe Fiameni}, {and} \bibinfo{person}{Rita
  Cucchiara}.} \bibinfo{year}{2022}\natexlab{}.
\newblock \showarticletitle{{Universal Captioner: Inducing Content-Style
  Separation in Vision-and-Language Model Training}}.
\newblock \bibinfo{journal}{\emph{arXiv preprint arXiv:2111.12727}}
  (\bibinfo{year}{2022}).
\newblock


\bibitem[Cornia et~al\mbox{.}(2020b)]%
        {cornia2020meshed}
\bibfield{author}{\bibinfo{person}{Marcella Cornia}, \bibinfo{person}{Matteo
  Stefanini}, \bibinfo{person}{Lorenzo Baraldi}, {and} \bibinfo{person}{Rita
  Cucchiara}.} \bibinfo{year}{2020}\natexlab{b}.
\newblock \showarticletitle{{Meshed-Memory Transformer for Image Captioning}}.
  In \bibinfo{booktitle}{\emph{CVPR}}.
\newblock


\bibitem[Devlin et~al\mbox{.}(2018)]%
        {devlin2018bert}
\bibfield{author}{\bibinfo{person}{Jacob Devlin}, \bibinfo{person}{Ming-Wei
  Chang}, \bibinfo{person}{Kenton Lee}, {and} \bibinfo{person}{Kristina
  Toutanova}.} \bibinfo{year}{2018}\natexlab{}.
\newblock \showarticletitle{{BERT: Pre-training of deep bidirectional
  transformers for language understanding}}. In
  \bibinfo{booktitle}{\emph{NAACL}}.
\newblock


\bibitem[Donahue et~al\mbox{.}(2015)]%
        {donahue2015long}
\bibfield{author}{\bibinfo{person}{Jeffrey Donahue}, \bibinfo{person}{Lisa
  Anne~Hendricks}, \bibinfo{person}{Sergio Guadarrama}, \bibinfo{person}{Marcus
  Rohrbach}, \bibinfo{person}{Subhashini Venugopalan}, \bibinfo{person}{Kate
  Saenko}, {and} \bibinfo{person}{Trevor Darrell}.}
  \bibinfo{year}{2015}\natexlab{}.
\newblock \showarticletitle{{Long-term recurrent convolutional networks for
  visual recognition and description}}. In \bibinfo{booktitle}{\emph{CVPR}}.
\newblock


\bibitem[Dosovitskiy et~al\mbox{.}(2021)]%
        {dosovitskiy2021image}
\bibfield{author}{\bibinfo{person}{Alexey Dosovitskiy}, \bibinfo{person}{Lucas
  Beyer}, \bibinfo{person}{Alexander Kolesnikov}, \bibinfo{person}{Dirk
  Weissenborn}, \bibinfo{person}{Xiaohua Zhai}, \bibinfo{person}{Thomas
  Unterthiner}, \bibinfo{person}{Mostafa Dehghani}, \bibinfo{person}{Matthias
  Minderer}, \bibinfo{person}{Georg Heigold}, \bibinfo{person}{Sylvain Gelly},
  \bibinfo{person}{Jakob Uszkoreit}, {and} \bibinfo{person}{Neil Houlsby}.}
  \bibinfo{year}{2021}\natexlab{}.
\newblock \showarticletitle{{An Image is Worth 16x16 Words: Transformers for
  Image Recognition at Scale}}. In \bibinfo{booktitle}{\emph{ICLR}}.
\newblock


\bibitem[Dubey et~al\mbox{.}(2021)]%
        {dubey2021vision}
\bibfield{author}{\bibinfo{person}{Shiv~Ram Dubey},
  \bibinfo{person}{Satish~Kumar Singh}, {and} \bibinfo{person}{Wei-Ta Chu}.}
  \bibinfo{year}{2021}\natexlab{}.
\newblock \showarticletitle{{Vision Transformer Hashing for Image Retrieval}}.
  In \bibinfo{booktitle}{\emph{ICME}}.
\newblock


\bibitem[El-Nouby et~al\mbox{.}(2021)]%
        {el2021training}
\bibfield{author}{\bibinfo{person}{Alaaeldin El-Nouby},
  \bibinfo{person}{Natalia Neverova}, \bibinfo{person}{Ivan Laptev}, {and}
  \bibinfo{person}{Herv{\'e} J{\'e}gou}.} \bibinfo{year}{2021}\natexlab{}.
\newblock \showarticletitle{{Training Vision Transformers for Image
  Retrieval}}.
\newblock \bibinfo{journal}{\emph{arXiv preprint arXiv:2102.05644}}
  (\bibinfo{year}{2021}).
\newblock


\bibitem[Guo et~al\mbox{.}(2020)]%
        {guo2020normalized}
\bibfield{author}{\bibinfo{person}{Longteng Guo}, \bibinfo{person}{Jing Liu},
  \bibinfo{person}{Xinxin Zhu}, \bibinfo{person}{Peng Yao},
  \bibinfo{person}{Shichen Lu}, {and} \bibinfo{person}{Hanqing Lu}.}
  \bibinfo{year}{2020}\natexlab{}.
\newblock \showarticletitle{{Normalized and Geometry-Aware Self-Attention
  Network for Image Captioning}}. In \bibinfo{booktitle}{\emph{CVPR}}.
\newblock


\bibitem[Guu et~al\mbox{.}(2020)]%
        {guu2020realm}
\bibfield{author}{\bibinfo{person}{Kelvin Guu}, \bibinfo{person}{Kenton Lee},
  \bibinfo{person}{Zora Tung}, \bibinfo{person}{Panupong Pasupat}, {and}
  \bibinfo{person}{Ming-Wei Chang}.} \bibinfo{year}{2020}\natexlab{}.
\newblock \showarticletitle{{REALM: Retrieval-Augmented Language Model
  Pre-Training}}. In \bibinfo{booktitle}{\emph{ICML}}.
\newblock


\bibitem[He et~al\mbox{.}(2016)]%
        {he2016deep}
\bibfield{author}{\bibinfo{person}{Kaiming He}, \bibinfo{person}{Xiangyu
  Zhang}, \bibinfo{person}{Shaoqing Ren}, {and} \bibinfo{person}{Jian Sun}.}
  \bibinfo{year}{2016}\natexlab{}.
\newblock \showarticletitle{Deep residual learning for image recognition}. In
  \bibinfo{booktitle}{\emph{CVPR}}.
\newblock


\bibitem[Herdade et~al\mbox{.}(2019)]%
        {herdade2019image}
\bibfield{author}{\bibinfo{person}{Simao Herdade}, \bibinfo{person}{Armin
  Kappeler}, \bibinfo{person}{Kofi Boakye}, {and} \bibinfo{person}{Joao
  Soares}.} \bibinfo{year}{2019}\natexlab{}.
\newblock \showarticletitle{{Image Captioning: Transforming Objects into
  Words}}. In \bibinfo{booktitle}{\emph{NeurIPS}}.
\newblock


\bibitem[Hu et~al\mbox{.}(2022)]%
        {hu2021scaling}
\bibfield{author}{\bibinfo{person}{Xiaowei Hu}, \bibinfo{person}{Zhe Gan},
  \bibinfo{person}{Jianfeng Wang}, \bibinfo{person}{Zhengyuan Yang},
  \bibinfo{person}{Zicheng Liu}, \bibinfo{person}{Yumao Lu}, {and}
  \bibinfo{person}{Lijuan Wang}.} \bibinfo{year}{2022}\natexlab{}.
\newblock \showarticletitle{{Scaling Up Vision-Language Pre-Training for Image
  Captioning}}. In \bibinfo{booktitle}{\emph{CVPR}}.
\newblock


\bibitem[Huang et~al\mbox{.}(2019)]%
        {huang2019attention}
\bibfield{author}{\bibinfo{person}{Lun Huang}, \bibinfo{person}{Wenmin Wang},
  \bibinfo{person}{Jie Chen}, {and} \bibinfo{person}{Xiao-Yong Wei}.}
  \bibinfo{year}{2019}\natexlab{}.
\newblock \showarticletitle{{Attention on Attention for Image Captioning}}. In
  \bibinfo{booktitle}{\emph{ICCV}}.
\newblock


\bibitem[Izacard and Grave(2021)]%
        {izacard2020leveraging}
\bibfield{author}{\bibinfo{person}{Gautier Izacard} {and}
  \bibinfo{person}{Edouard Grave}.} \bibinfo{year}{2021}\natexlab{}.
\newblock \showarticletitle{{Leveraging Passage Retrieval with Generative
  Models for Open Domain Question Answering}}. In
  \bibinfo{booktitle}{\emph{EACL}}.
\newblock


\bibitem[Johnson et~al\mbox{.}(2019)]%
        {johnson2019billion}
\bibfield{author}{\bibinfo{person}{Jeff Johnson}, \bibinfo{person}{Matthijs
  Douze}, {and} \bibinfo{person}{Herv{\'e} J{\'e}gou}.}
  \bibinfo{year}{2019}\natexlab{}.
\newblock \showarticletitle{Billion-scale similarity search with gpus}.
\newblock \bibinfo{journal}{\emph{IEEE Trans. Big Data}} \bibinfo{volume}{7},
  \bibinfo{number}{3} (\bibinfo{year}{2019}), \bibinfo{pages}{535--547}.
\newblock


\bibitem[Kaplan et~al\mbox{.}(2020)]%
        {kaplan2020scaling}
\bibfield{author}{\bibinfo{person}{Jared Kaplan}, \bibinfo{person}{Sam
  McCandlish}, \bibinfo{person}{Tom Henighan}, \bibinfo{person}{Tom~B Brown},
  \bibinfo{person}{Benjamin Chess}, \bibinfo{person}{Rewon Child},
  \bibinfo{person}{Scott Gray}, \bibinfo{person}{Alec Radford},
  \bibinfo{person}{Jeffrey Wu}, {and} \bibinfo{person}{Dario Amodei}.}
  \bibinfo{year}{2020}\natexlab{}.
\newblock \showarticletitle{Scaling laws for neural language models}.
\newblock \bibinfo{journal}{\emph{arXiv preprint arXiv:2001.08361}}
  (\bibinfo{year}{2020}).
\newblock


\bibitem[Karpathy and Fei-Fei(2015)]%
        {karpathy2015deep}
\bibfield{author}{\bibinfo{person}{Andrej Karpathy} {and} \bibinfo{person}{Li
  Fei-Fei}.} \bibinfo{year}{2015}\natexlab{}.
\newblock \showarticletitle{Deep visual-semantic alignments for generating
  image descriptions}. In \bibinfo{booktitle}{\emph{CVPR}}.
\newblock


\bibitem[Khandelwal et~al\mbox{.}(2020)]%
        {khandelwal2019generalization}
\bibfield{author}{\bibinfo{person}{Urvashi Khandelwal}, \bibinfo{person}{Omer
  Levy}, \bibinfo{person}{Dan Jurafsky}, \bibinfo{person}{Luke Zettlemoyer},
  {and} \bibinfo{person}{Mike Lewis}.} \bibinfo{year}{2020}\natexlab{}.
\newblock \showarticletitle{{Generalization through Memorization: Nearest
  Neighbor Language Models}}. In \bibinfo{booktitle}{\emph{ICLR}}.
\newblock


\bibitem[Kingma and Ba(2015)]%
        {kingma2015adam}
\bibfield{author}{\bibinfo{person}{Diederik~P Kingma} {and}
  \bibinfo{person}{Jimmy Ba}.} \bibinfo{year}{2015}\natexlab{}.
\newblock \showarticletitle{{Adam: A Method for Stochastic Optimization}}. In
  \bibinfo{booktitle}{\emph{ICLR}}.
\newblock


\bibitem[Landi et~al\mbox{.}(2021)]%
        {landi2021working}
\bibfield{author}{\bibinfo{person}{Federico Landi}, \bibinfo{person}{Lorenzo
  Baraldi}, \bibinfo{person}{Marcella Cornia}, {and} \bibinfo{person}{Rita
  Cucchiara}.} \bibinfo{year}{2021}\natexlab{}.
\newblock \showarticletitle{{Working Memory Connections for LSTM}}.
\newblock \bibinfo{journal}{\emph{Neural Networks}}  \bibinfo{volume}{144}
  (\bibinfo{year}{2021}), \bibinfo{pages}{334--341}.
\newblock


\bibitem[Lewis et~al\mbox{.}(2020)]%
        {lewis2020retrieval}
\bibfield{author}{\bibinfo{person}{Patrick Lewis}, \bibinfo{person}{Ethan
  Perez}, \bibinfo{person}{Aleksandra Piktus}, \bibinfo{person}{Fabio Petroni},
  \bibinfo{person}{Vladimir Karpukhin}, \bibinfo{person}{Naman Goyal},
  \bibinfo{person}{Heinrich K{\"u}ttler}, \bibinfo{person}{Mike Lewis},
  \bibinfo{person}{Wen-tau Yih}, \bibinfo{person}{Tim Rockt\"{a}schel},
  \bibinfo{person}{Sebastian Riedel}, {and} \bibinfo{person}{Douwe Kiela}.}
  \bibinfo{year}{2020}\natexlab{}.
\newblock \showarticletitle{{Retrieval-Augmented Generation for
  Knowledge-Intensive NLP Tasks}}. In \bibinfo{booktitle}{\emph{NeurIPS}}.
\newblock


\bibitem[Li et~al\mbox{.}(2020)]%
        {li2020oscar}
\bibfield{author}{\bibinfo{person}{Xiujun Li}, \bibinfo{person}{Xi Yin},
  \bibinfo{person}{Chunyuan Li}, \bibinfo{person}{Pengchuan Zhang},
  \bibinfo{person}{Xiaowei Hu}, \bibinfo{person}{Lei Zhang},
  \bibinfo{person}{Lijuan Wang}, \bibinfo{person}{Houdong Hu},
  \bibinfo{person}{Li Dong}, \bibinfo{person}{Furu Wei}, {et~al\mbox{.}}}
  \bibinfo{year}{2020}\natexlab{}.
\newblock \showarticletitle{{Oscar: Object-semantics aligned pre-training for
  vision-language tasks}}. In \bibinfo{booktitle}{\emph{ECCV}}.
\newblock


\bibitem[Lin(2004)]%
        {lin2004rouge}
\bibfield{author}{\bibinfo{person}{Chin-Yew Lin}.}
  \bibinfo{year}{2004}\natexlab{}.
\newblock \showarticletitle{Rouge: A package for automatic evaluation of
  summaries}. In \bibinfo{booktitle}{\emph{ACL Workshops}}.
\newblock


\bibitem[Lin et~al\mbox{.}(2014)]%
        {lin2014microsoft}
\bibfield{author}{\bibinfo{person}{Tsung-Yi Lin}, \bibinfo{person}{Michael
  Maire}, \bibinfo{person}{Serge Belongie}, \bibinfo{person}{James Hays},
  \bibinfo{person}{Pietro Perona}, \bibinfo{person}{Deva Ramanan},
  \bibinfo{person}{Piotr Doll{\'a}r}, {and} \bibinfo{person}{C~Lawrence
  Zitnick}.} \bibinfo{year}{2014}\natexlab{}.
\newblock \showarticletitle{{Microsoft COCO: Common Objects in Context}}. In
  \bibinfo{booktitle}{\emph{ECCV}}.
\newblock


\bibitem[Liu et~al\mbox{.}(2020)]%
        {liu2020prophet}
\bibfield{author}{\bibinfo{person}{Fenglin Liu}, \bibinfo{person}{Xuancheng
  Ren}, \bibinfo{person}{Xian Wu}, \bibinfo{person}{Shen Ge},
  \bibinfo{person}{Wei Fan}, \bibinfo{person}{Yuexian Zou}, {and}
  \bibinfo{person}{Xu Sun}.} \bibinfo{year}{2020}\natexlab{}.
\newblock \showarticletitle{{Prophet Attention: Predicting Attention with
  Future Attention}}. In \bibinfo{booktitle}{\emph{NeurIPS}}.
\newblock


\bibitem[Liu et~al\mbox{.}(2021)]%
        {liu2021cptr}
\bibfield{author}{\bibinfo{person}{Wei Liu}, \bibinfo{person}{Sihan Chen},
  \bibinfo{person}{Longteng Guo}, \bibinfo{person}{Xinxin Zhu}, {and}
  \bibinfo{person}{Jing Liu}.} \bibinfo{year}{2021}\natexlab{}.
\newblock \showarticletitle{{CPTR: Full Transformer Network for Image
  Captioning}}.
\newblock \bibinfo{journal}{\emph{arXiv preprint arXiv:2101.10804}}
  (\bibinfo{year}{2021}).
\newblock


\bibitem[Luo et~al\mbox{.}(2021)]%
        {luo2021dual}
\bibfield{author}{\bibinfo{person}{Yunpeng Luo}, \bibinfo{person}{Jiayi Ji},
  \bibinfo{person}{Xiaoshuai Sun}, \bibinfo{person}{Liujuan Cao},
  \bibinfo{person}{Yongjian Wu}, \bibinfo{person}{Feiyue Huang},
  \bibinfo{person}{Chia-Wen Lin}, {and} \bibinfo{person}{Rongrong Ji}.}
  \bibinfo{year}{2021}\natexlab{}.
\newblock \showarticletitle{{Dual-Level Collaborative Transformer for Image
  Captioning}}. In \bibinfo{booktitle}{\emph{AAAI}}.
\newblock


\bibitem[Micikevicius et~al\mbox{.}(2018)]%
        {micikevicius2017mixed}
\bibfield{author}{\bibinfo{person}{Paulius Micikevicius},
  \bibinfo{person}{Sharan Narang}, \bibinfo{person}{Jonah Alben},
  \bibinfo{person}{Gregory Diamos}, \bibinfo{person}{Erich Elsen},
  \bibinfo{person}{David Garcia}, \bibinfo{person}{Boris Ginsburg},
  \bibinfo{person}{Michael Houston}, \bibinfo{person}{Oleksii Kuchaiev},
  \bibinfo{person}{Ganesh Venkatesh}, {and} \bibinfo{person}{Hao Wu}.}
  \bibinfo{year}{2018}\natexlab{}.
\newblock \showarticletitle{{Mixed Precision Training}}. In
  \bibinfo{booktitle}{\emph{ICLR}}.
\newblock


\bibitem[Pan et~al\mbox{.}(2020)]%
        {pan2020x}
\bibfield{author}{\bibinfo{person}{Yingwei Pan}, \bibinfo{person}{Ting Yao},
  \bibinfo{person}{Yehao Li}, {and} \bibinfo{person}{Tao Mei}.}
  \bibinfo{year}{2020}\natexlab{}.
\newblock \showarticletitle{{X-Linear Attention Networks for Image
  Captioning}}. In \bibinfo{booktitle}{\emph{CVPR}}.
\newblock


\bibitem[Papineni et~al\mbox{.}(2002)]%
        {papineni2002bleu}
\bibfield{author}{\bibinfo{person}{Kishore Papineni}, \bibinfo{person}{Salim
  Roukos}, \bibinfo{person}{Todd Ward}, {and} \bibinfo{person}{Wei-Jing Zhu}.}
  \bibinfo{year}{2002}\natexlab{}.
\newblock \showarticletitle{{BLEU: a method for automatic evaluation of machine
  translation}}. In \bibinfo{booktitle}{\emph{ACL}}.
\newblock


\bibitem[Radford et~al\mbox{.}(2021)]%
        {radford2021learning}
\bibfield{author}{\bibinfo{person}{Alec Radford}, \bibinfo{person}{Jong~Wook
  Kim}, \bibinfo{person}{Chris Hallacy}, \bibinfo{person}{Aditya Ramesh},
  \bibinfo{person}{Gabriel Goh}, \bibinfo{person}{Sandhini Agarwal},
  \bibinfo{person}{Girish Sastry}, \bibinfo{person}{Amanda Askell},
  \bibinfo{person}{Pamela Mishkin}, \bibinfo{person}{Jack Clark},
  \bibinfo{person}{Gretchen Krueger}, {and} \bibinfo{person}{Ilya Sutskever}.}
  \bibinfo{year}{2021}\natexlab{}.
\newblock \showarticletitle{{Learning Transferable Visual Models From Natural
  Language Supervision}}. In \bibinfo{booktitle}{\emph{ICML}}.
\newblock


\bibitem[Radford et~al\mbox{.}(2019)]%
        {Radford2019LanguageMA}
\bibfield{author}{\bibinfo{person}{Alec Radford}, \bibinfo{person}{Jeffrey Wu},
  \bibinfo{person}{Rewon Child}, \bibinfo{person}{David Luan},
  \bibinfo{person}{Dario Amodei}, {and} \bibinfo{person}{Ilya Sutskever}.}
  \bibinfo{year}{2019}\natexlab{}.
\newblock \showarticletitle{{Language Models are Unsupervised Multitask
  Learners}}.
\newblock \bibinfo{journal}{\emph{OpenAI Blog}} \bibinfo{volume}{1},
  \bibinfo{number}{8} (\bibinfo{year}{2019}), \bibinfo{pages}{9}.
\newblock


\bibitem[Rajbhandari et~al\mbox{.}(2020)]%
        {rajbhandari2020zero}
\bibfield{author}{\bibinfo{person}{Samyam Rajbhandari}, \bibinfo{person}{Jeff
  Rasley}, \bibinfo{person}{Olatunji Ruwase}, {and} \bibinfo{person}{Yuxiong
  He}.} \bibinfo{year}{2020}\natexlab{}.
\newblock \showarticletitle{{ZeRO: Memory optimizations Toward Training
  Trillion Parameter Models}}. In \bibinfo{booktitle}{\emph{SC}}.
\newblock


\bibitem[Rennie et~al\mbox{.}(2017)]%
        {rennie2017self}
\bibfield{author}{\bibinfo{person}{Steven~J Rennie}, \bibinfo{person}{Etienne
  Marcheret}, \bibinfo{person}{Youssef Mroueh}, \bibinfo{person}{Jarret Ross},
  {and} \bibinfo{person}{Vaibhava Goel}.} \bibinfo{year}{2017}\natexlab{}.
\newblock \showarticletitle{Self-critical sequence training for image
  captioning}. In \bibinfo{booktitle}{\emph{CVPR}}.
\newblock


\bibitem[Sennrich et~al\mbox{.}(2016)]%
        {sennrich2016neural}
\bibfield{author}{\bibinfo{person}{Rico Sennrich}, \bibinfo{person}{Barry
  Haddow}, {and} \bibinfo{person}{Alexandra Birch}.}
  \bibinfo{year}{2016}\natexlab{}.
\newblock \showarticletitle{{Neural Machine Translation of Rare Words with
  Subword Units}}. In \bibinfo{booktitle}{\emph{ACL}}.
\newblock


\bibitem[Shen et~al\mbox{.}(2021)]%
        {shen2021much}
\bibfield{author}{\bibinfo{person}{Sheng Shen}, \bibinfo{person}{Liunian~Harold
  Li}, \bibinfo{person}{Hao Tan}, \bibinfo{person}{Mohit Bansal},
  \bibinfo{person}{Anna Rohrbach}, \bibinfo{person}{Kai-Wei Chang},
  \bibinfo{person}{Zhewei Yao}, {and} \bibinfo{person}{Kurt Keutzer}.}
  \bibinfo{year}{2021}\natexlab{}.
\newblock \showarticletitle{{How Much Can CLIP Benefit Vision-and-Language
  Tasks?}}
\newblock \bibinfo{journal}{\emph{arXiv preprint arXiv:2107.06383}}
  (\bibinfo{year}{2021}).
\newblock


\bibitem[Shi et~al\mbox{.}(2020)]%
        {shi2020improving}
\bibfield{author}{\bibinfo{person}{Zhan Shi}, \bibinfo{person}{Xu Zhou},
  \bibinfo{person}{Xipeng Qiu}, {and} \bibinfo{person}{Xiaodan Zhu}.}
  \bibinfo{year}{2020}\natexlab{}.
\newblock \showarticletitle{{Improving Image Captioning with Better Use of
  Captions}}. In \bibinfo{booktitle}{\emph{ACL}}.
\newblock


\bibitem[Sukhbaatar et~al\mbox{.}(2019)]%
        {sukhbaatar2019augmenting}
\bibfield{author}{\bibinfo{person}{Sainbayar Sukhbaatar},
  \bibinfo{person}{Edouard Grave}, \bibinfo{person}{Guillaume Lample},
  \bibinfo{person}{Herve Jegou}, {and} \bibinfo{person}{Armand Joulin}.}
  \bibinfo{year}{2019}\natexlab{}.
\newblock \showarticletitle{{Augmenting Self-Attention with Persistent
  Memory}}.
\newblock \bibinfo{journal}{\emph{arXiv preprint arXiv:1907.01470}}
  (\bibinfo{year}{2019}).
\newblock


\bibitem[Tolias et~al\mbox{.}(2016)]%
        {tolias2016particular}
\bibfield{author}{\bibinfo{person}{Giorgos Tolias}, \bibinfo{person}{Ronan
  Sicre}, {and} \bibinfo{person}{Herv{\'e} J{\'e}gou}.}
  \bibinfo{year}{2016}\natexlab{}.
\newblock \showarticletitle{Particular object retrieval with integral
  max-pooling of CNN activations}. In \bibinfo{booktitle}{\emph{ICLR}}.
\newblock


\bibitem[Touvron et~al\mbox{.}(2021)]%
        {touvron2021training}
\bibfield{author}{\bibinfo{person}{Hugo Touvron}, \bibinfo{person}{Matthieu
  Cord}, \bibinfo{person}{Matthijs Douze}, \bibinfo{person}{Francisco Massa},
  \bibinfo{person}{Alexandre Sablayrolles}, {and} \bibinfo{person}{Herv{\'e}
  J{\'e}gou}.} \bibinfo{year}{2021}\natexlab{}.
\newblock \showarticletitle{{Training data-efficient image transformers \&
  distillation through attention}}. In \bibinfo{booktitle}{\emph{ICML}}.
\newblock


\bibitem[Vaswani et~al\mbox{.}(2017)]%
        {vaswani2017attention}
\bibfield{author}{\bibinfo{person}{Ashish Vaswani}, \bibinfo{person}{Noam
  Shazeer}, \bibinfo{person}{Niki Parmar}, \bibinfo{person}{Jakob Uszkoreit},
  \bibinfo{person}{Llion Jones}, \bibinfo{person}{Aidan~N Gomez},
  \bibinfo{person}{{\L}ukasz Kaiser}, {and} \bibinfo{person}{Illia
  Polosukhin}.} \bibinfo{year}{2017}\natexlab{}.
\newblock \showarticletitle{Attention is all you need}. In
  \bibinfo{booktitle}{\emph{NeurIPS}}.
\newblock


\bibitem[Vedantam et~al\mbox{.}(2015)]%
        {vedantam2015cider}
\bibfield{author}{\bibinfo{person}{Ramakrishna Vedantam}, \bibinfo{person}{C
  Lawrence~Zitnick}, {and} \bibinfo{person}{Devi Parikh}.}
  \bibinfo{year}{2015}\natexlab{}.
\newblock \showarticletitle{{CIDEr: Consensus-based Image Description
  Evaluation}}. In \bibinfo{booktitle}{\emph{CVPR}}.
\newblock


\bibitem[Vinyals et~al\mbox{.}(2015)]%
        {vinyals2015show}
\bibfield{author}{\bibinfo{person}{Oriol Vinyals}, \bibinfo{person}{Alexander
  Toshev}, \bibinfo{person}{Samy Bengio}, {and} \bibinfo{person}{Dumitru
  Erhan}.} \bibinfo{year}{2015}\natexlab{}.
\newblock \showarticletitle{Show and tell: A neural image caption generator}.
  In \bibinfo{booktitle}{\emph{CVPR}}.
\newblock


\bibitem[Wu et~al\mbox{.}(2022)]%
        {wu2022memorizing}
\bibfield{author}{\bibinfo{person}{Yuhuai Wu}, \bibinfo{person}{Markus~N Rabe},
  \bibinfo{person}{DeLesley Hutchins}, {and} \bibinfo{person}{Christian
  Szegedy}.} \bibinfo{year}{2022}\natexlab{}.
\newblock \showarticletitle{{Memorizing Transformers}}. In
  \bibinfo{booktitle}{\emph{ICLR}}.
\newblock


\bibitem[Yang et~al\mbox{.}(2019a)]%
        {yang2019auto}
\bibfield{author}{\bibinfo{person}{Xu Yang}, \bibinfo{person}{Kaihua Tang},
  \bibinfo{person}{Hanwang Zhang}, {and} \bibinfo{person}{Jianfei Cai}.}
  \bibinfo{year}{2019}\natexlab{a}.
\newblock \showarticletitle{{Auto-Encoding Scene Graphs for Image Captioning}}.
  In \bibinfo{booktitle}{\emph{CVPR}}.
\newblock


\bibitem[Yang et~al\mbox{.}(2019b)]%
        {yang2019learning}
\bibfield{author}{\bibinfo{person}{Xu Yang}, \bibinfo{person}{Hanwang Zhang},
  {and} \bibinfo{person}{Jianfei Cai}.} \bibinfo{year}{2019}\natexlab{b}.
\newblock \showarticletitle{{Learning to Collocate Neural Modules for Image
  Captioning}}. In \bibinfo{booktitle}{\emph{ICCV}}.
\newblock


\bibitem[Yao et~al\mbox{.}(2018)]%
        {yao2018exploring}
\bibfield{author}{\bibinfo{person}{Ting Yao}, \bibinfo{person}{Yingwei Pan},
  \bibinfo{person}{Yehao Li}, {and} \bibinfo{person}{Tao Mei}.}
  \bibinfo{year}{2018}\natexlab{}.
\newblock \showarticletitle{{Exploring Visual Relationship for Image
  Captioning}}. In \bibinfo{booktitle}{\emph{ECCV}}.
\newblock


\bibitem[Yogatama et~al\mbox{.}(2021)]%
        {yogatama2021adaptive}
\bibfield{author}{\bibinfo{person}{Dani Yogatama}, \bibinfo{person}{Cyprien de
  Masson~d’Autume}, {and} \bibinfo{person}{Lingpeng Kong}.}
  \bibinfo{year}{2021}\natexlab{}.
\newblock \showarticletitle{Adaptive Semiparametric Language Models}.
\newblock \bibinfo{journal}{\emph{TACL}}  \bibinfo{volume}{9}
  (\bibinfo{year}{2021}), \bibinfo{pages}{362--373}.
\newblock


\bibitem[You et~al\mbox{.}(2020)]%
        {you2019large}
\bibfield{author}{\bibinfo{person}{Yang You}, \bibinfo{person}{Jing Li},
  \bibinfo{person}{Sashank Reddi}, \bibinfo{person}{Jonathan Hseu},
  \bibinfo{person}{Sanjiv Kumar}, \bibinfo{person}{Srinadh Bhojanapalli},
  \bibinfo{person}{Xiaodan Song}, \bibinfo{person}{James Demmel},
  \bibinfo{person}{Kurt Keutzer}, {and} \bibinfo{person}{Cho-Jui Hsieh}.}
  \bibinfo{year}{2020}\natexlab{}.
\newblock \showarticletitle{{Large Batch Optimization for Deep Learning:
  Training BERT in 76 minutes}}. In \bibinfo{booktitle}{\emph{ICLR}}.
\newblock


\bibitem[Zhang et~al\mbox{.}(2021)]%
        {zhang2021rstnet}
\bibfield{author}{\bibinfo{person}{Xuying Zhang}, \bibinfo{person}{Xiaoshuai
  Sun}, \bibinfo{person}{Yunpeng Luo}, \bibinfo{person}{Jiayi Ji},
  \bibinfo{person}{Yiyi Zhou}, \bibinfo{person}{Yongjian Wu},
  \bibinfo{person}{Feiyue Huang}, {and} \bibinfo{person}{Rongrong Ji}.}
  \bibinfo{year}{2021}\natexlab{}.
\newblock \showarticletitle{{RSTNet: Captioning with Adaptive Attention on
  Visual and Non-Visual Words}}. In \bibinfo{booktitle}{\emph{CVPR}}.
\newblock


\bibitem[Zhou et~al\mbox{.}(2020)]%
        {zhou2020unified}
\bibfield{author}{\bibinfo{person}{Luowei Zhou}, \bibinfo{person}{Hamid
  Palangi}, \bibinfo{person}{Lei Zhang}, \bibinfo{person}{Houdong Hu},
  \bibinfo{person}{Jason~J Corso}, {and} \bibinfo{person}{Jianfeng Gao}.}
  \bibinfo{year}{2020}\natexlab{}.
\newblock \showarticletitle{{Unified Vision-Language Pre-Training for Image
  Captioning and VQA}}. In \bibinfo{booktitle}{\emph{AAAI}}.
\newblock


\end{thebibliography}


\end{document}